\documentclass[10pt,twocolumn,letterpaper]{article}

\usepackage{iccv}
\usepackage{times}
\usepackage{epsfig}
\usepackage{graphicx}
\usepackage{amsmath}
\usepackage{amssymb}
\usepackage{multirow}
\usepackage{bbding}
\usepackage{bbm}
\usepackage{bm}
\usepackage{listings}
\usepackage{subfigure} 
\usepackage{booktabs}
\usepackage{pifont} 
\newcommand*\colourcheck[1]{%
  \expandafter\newcommand\csname #1check\endcsname{\textcolor{#1}{\ding{52}}}%
}
\colourcheck{blue}
\colourcheck{green}
\colourcheck{red}
\newcommand*\colourtime[1]{%
  \expandafter\newcommand\csname #1time\endcsname{\textcolor{#1}{\ding{56}}}%
}
\colourtime{blue}
\colourtime{green}
\colourtime{red}

\usepackage[pagebackref=true,breaklinks=true,letterpaper=true,colorlinks,bookmarks=false]{hyperref}

\iccvfinalcopy %

\def\iccvPaperID{1883} %

\ificcvfinal\fi

\begin{document}

\title{Learning to Estimate 6DoF Pose from Limited Data: \\ A Few-Shot, Generalizable Approach using RGB Images}

\author{
  Panwang Pan\textsuperscript{1}\thanks{Equal contribution},
  Zhiwen Fan\textsuperscript{2$\ast$},
  Brandon Y. Feng\textsuperscript{3$\ast$},
  Peihao Wang\textsuperscript{2$\ast$},
  Chenxin Li\textsuperscript{4},
  Zhangyang Wang\textsuperscript{2}\\
  {\textsuperscript{1}ByteDance}, \, {\textsuperscript{2}University of Texas at Austin},\\ {\textsuperscript{3}University of Maryland}, \,{\textsuperscript{4}Hong Kong Polytechnic University}\\
}

\maketitle
\ificcvfinal\thispagestyle{empty}\fi

\begin{abstract}
The accurate estimation of six degrees-of-freedom (6DoF) object poses is essential for many applications in robotics and augmented reality. However, existing methods for 6DoF pose estimation often depend on CAD templates or dense support views, restricting their usefulness in real-world situations. In this study, we present a new cascade framework named \textbf{Cas6D} for few-shot 6DoF pose estimation that is generalizable and uses only RGB images.

To address the false positives of target object detection in the extreme few-shot setting, our framework utilizes a self-supervised pre-trained ViT to learn robust feature representations. Then, we initialize the nearest top-K pose candidates based on similarity score and refine the initial poses using feature pyramids to formulate and update the cascade warped feature volume, which encodes context at increasingly finer scales. By discretizing the pose search range using multiple pose bins and progressively narrowing the pose search range in each stage using predictions from the previous stage, Cas6D can overcome the large gap between pose candidates and ground truth poses, which is a common failure mode in sparse-view scenarios.
Experimental results on the LINEMOD and GenMOP datasets demonstrate that Cas6D outperforms state-of-the-art methods by 9.2\% and 3.8\% accuracy (Proj-5) under the 32-shot setting compared to OnePose++ and Gen6D. Our framework also performs best under the full-shot setting with all support views. The code will be released.

\end{abstract}

\section{Introduction}
\begin{figure}[!ht]\label{fig:teaser}
    \centering
    \includegraphics[width=0.99\linewidth]{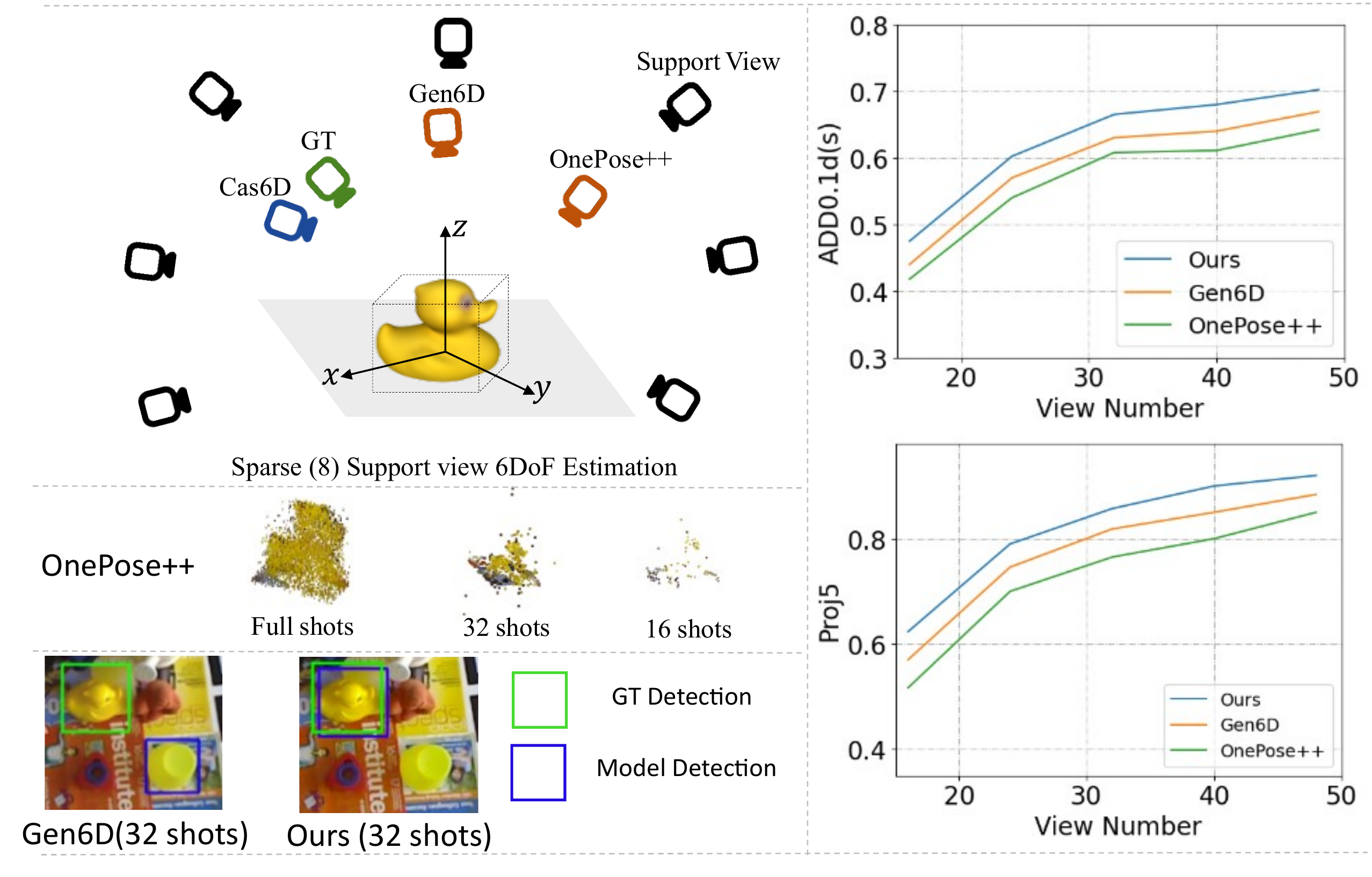}
    \caption{\textbf{Generalizable 6DoF Estimation from Sparse RGB Views without CAD Models.} While OnePose++~\cite{he2023onepose++} (middle left) and Gen6D~\cite{gen6d} (bottom left) achieve promising accuracy given sufficient support views, their performance significantly degrades when few views are available. In contrast, Cas6D bridges the performance gap (+3.8\% on Prj-5) on few-shot settings (right) without the need for depth maps or per-instance re-training.}
    \vspace{-4mm}
\end{figure}
Six degrees-of-freedom (6DoF) object pose estimation predicts the orientation and location of a target object in 3D space, which is a necessary preliminary step for downstream applications like robotic manipulation, augmented reality, and autonomous driving. 
However, traditional methods~\cite{kehl2017ssd,tekin2018real,xiang2017posecnn,zakharov2019dpod,li2019cdpn,peng2019pvnet,labbe2020cosypose} often require instance-level CAD models as prior knowledge, which can be difficult or impossible to obtain for objects in the wild. While category-level methods~\cite{wang2019normalized,ahmadyan2021objectron,chen2020category} eliminate the need for instance-level CAD models, they are limited to handling different instances within the same category and cannot generalize to unseen categories.
To overcome these limitations, generalizable (model-free) methods, OnePose~\cite{sun2022onepose} and OnePose++~\cite{he2023onepose++} propose a one-shot object pose estimation method that can be used in arbitrary scenarios, which first reconstructs sparse object point clouds and then establishes 2D-3D correspondences between keypoints in the query image and the point cloud to estimate the object pose. Another recent method, Gen6D~\cite{gen6d}, uses a set of support views to detect the target object in a query image, selects a support view with the most similar appearance to provide the initial pose, and then refines the pose estimation using a feature volume constructed from 2D features.

\begin{figure*}[!t] \label{fig:main_arc}
    \centering
    \includegraphics[width=0.9\linewidth]{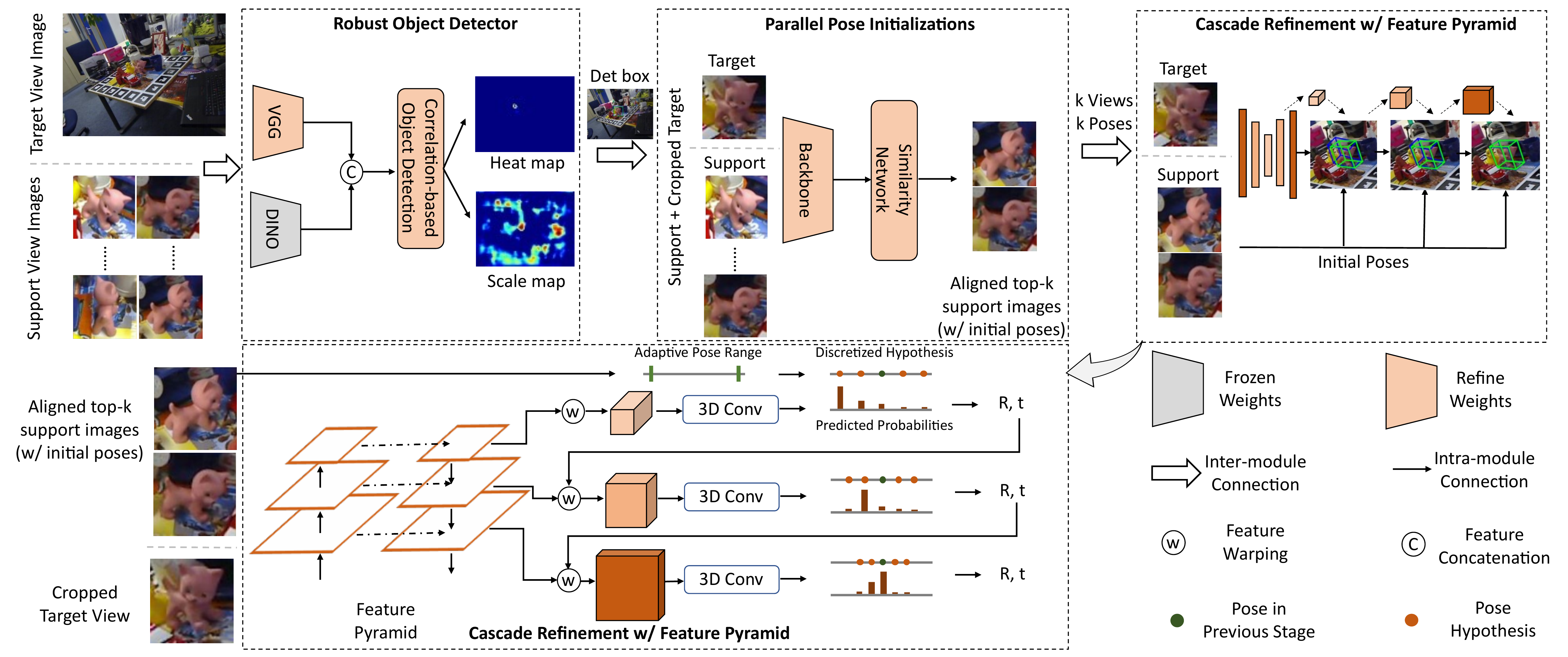} 
    \caption{\textbf{The Overall Pipeline of the Proposed Cas6D Framework.} Given multi-view object images in support views and a target view image, our pipeline consists of three stages. Firstly, we detect and crop the object by localizing the heat and scale maps (top left). Secondly, we use a similarity network to find the best top-K aligned support images as pose initializations. Finally, we use the cascade pose refiner with feature pyramid to progressively estimate the pose residual by narrowing the search range.}

\end{figure*}

Previous methods have shown good generalizability to new instances, but require dense support views at inference(e.g., $\geq$ 128) which limits their effectiveness in real-world applications where images are often non-uniform and sparse~\cite{niemeyer2022regnerf}.
This can result in a failure to reconstruct the point cloud with sparse viewpoints(e.g., OnePose++~\cite{he2023onepose++}) or detect an accurate box(e.g., Gen6D~\cite{gen6d}), leading to a large gap between the initial pose and ground truth poses, which are illustrated in Figure~\ref{fig:teaser}.
While recent work, FS6D~\cite{he2022fs6d}, has introduced a large-scale synthetic dataset to address this issue, it still requires depth maps as input, which is a significant constraint for real-world applications.

Can we design an efficient RGB-based 6DoF pose estimator without requiring additional 3D information (e.g., CAD model, depth map) under sparse support views?
In this paper, we propose Cas6D, a novel CAD model-free solution for efficient RGB-based 6DoF pose estimation under sparse support views.
Our approach leverages features extracted by the self-supervised ViT representation~\cite{caron2021emerging} to identify the correspondences for object parts~\cite{amir2021deep} across views and thus detect the object in the target view. Then, we use a similarity network to find the top-K best-matched support images as pose initializations.
To bridge the large gap between the initial poses and the sparse ground truth, we propose a cascaded coarse-to-fine refinement module. 
During training, we determine the initial pose gap and sample multiple pose hypotheses using a low-resolution feature volume based on coarse but semantic-rich 2D features.
In later stages, we utilize pose estimates from earlier stages to adjust the sampling range of pose hypotheses and construct feature volumes of higher resolution with finer 2D semantic features. This framework allows for adaptive pose sampling with adjustable feature pyramids, which avoids using a single-scale 3D volume and ensures computational and memory resource efficiency.

Experiments on the LINEMOD~\cite{hinterstoisser2013model} and GenMOP~\cite{gen6d} datasets, show that Cas6D consistently outperforms other RGB-based generalizable and instance-wise methods.
We summarize the contributions as follows:
\begin{itemize}
    \item We address the challenging setting of few-shot, generalizable 6D object pose estimation using only RGB images. Our proposed Cas6D framework progressively estimates the pose of the target view from a limited number of support views.
    \item Our framework leverages representations extracted from a self-supervised pre-trained ViT network and employs a top-K pose proposals scheme for robust pose initialization. It also introduces a cascaded coarse-to-fine refinement procedure that uses feature pyramids and adaptively discrete pose hypotheses.
    \item We extensively evaluate our method on the widely used LINEMOD and GenMOP benchmarks, demonstrating that Cas6D outperforms previous state-of-the-art methods OnePose++ by  9.2\%/4.7\% and Gen6D by 3.8\%/2.8\%, measured by ADD-0.1d/Prj-5 metrics.
\end{itemize}

\label{sec:intro}

\section{Related Works}
\paragraph{Generalizable Object Pose Estimator.}
Recent advances in generalizable pose estimators aim to address the limitations of prior instance-specific methods~\cite{xiang2017posecnn,sundermeyer2018implicit,hodan2018bop,peng2019pvnet,labbe2020cosypose,hodavn2020bop,hodan2020epos,hu2020single,labbe2020cosypose,wen2020edge,di2021so,liu2020keypose,song2020hybridpose,ponimatkin2022focal,su2022zebrapose,yen2021inerf} which typically utilize CAD models or depth maps
and category-specific~\cite{wang2019normalized,chen2020category,wen2021disentangled,deng2022icaps,lin2021sparse,chen2020category,chen2021fs,lin2021dualposenet,tian2020shape,di2022gpv,goodwin2022zero} 
methods that require per-category training, enabling pose estimation generalizing to unseen objects and categories. 
Depending on whether a 3D model is required, 
existing literature on generalizable pose estimation could be divided into two streams: model-based and model-free. The former requires high-quality object models either for shape embedding~\cite{xiao2019pose,pitteri20203d,dani20213dposelite}, template matching~\cite{hinterstoisser2011multimodal,balntas2017pose,wohlhart2015learning,sundermeyer2020multi}, rendering-and-comparison approaches~\cite{li2018deepim, zakharov2019dpod, okorn2021zephyr, busam2020like}.
To avoid the reliance on 3D models, the latter model-free methods utilize advanced neural rendering to directly render from posed images for pose estimation~\cite{yen2020inerf,park2020latentfusion,lin2022parallel}. But some methods still assume the availability of depth maps~\cite{park2020latentfusion} or object masks~\cite{yen2020inerf,park2020latentfusion,lin2022parallel}. Besides, the iterative optimization process~\cite{yen2020inerf} is time-consuming and the work~\cite{lin2022parallel} introduce instant-ngp~\cite{muller2022instant} as neural renderer with parallelized Monte Carlo sampling to mitigate the speed limitation.
Other model-free methods such as Gen6D~\cite{gen6d}, OnePose~\cite{sun2022onepose} and OnePose++~\cite{he2023onepose++}, use only a set of reference/support images with annotated poses for the pose estimation of the object in the target image. Specifically, Gen6D~\cite{gen6d} first detects the object box in the target view, initializes a pose from dense support views and refines the pose using feature volume and multiple 3D convolution layers. OnePose series~\cite{sun2022onepose,he2023onepose++} first reconstruct the sparse point cloud from the RGB sequences of all support viewpoints, and match the target view with the sparse point cloud to determine the object poses.
However, the above method easily fails under sparse view scenarios as Gen6D~\cite{gen6d} may detect the incorrect box and fail to refine the pose due to large gap between initial and GT poses. OnePose series~\cite{sun2022onepose,he2023onepose++} are not able to build high-quality point clouds due to sparse support view.
As a comparison, we leverage the rich semantic prior from pre-trained ViT models (e.g., DINO-ViT~\cite{caron2021emerging}) for the accurate detection of an arbitrary object. We initialize multiple parallel poses from sparse support views based on the similarity scores, and progressively refine the object pose with cascade feature pyramid volumes and discrete pose hypotheses.
\vspace{-5mm}
\paragraph{Few-shot Learning in 3D Vision.}
Few-shot learning is to develop a model that can generalize to new classes with very few examples, through meta-learning approaches~\cite{finn2017model,garcia2017few,munkhdalai2017meta,ravi2017optimization} or metric-based fashion~\cite{garcia2017few,snell2017prototypical,vinyals2016matching}. 
For 6D pose estimation, the few-shot pose estimator usually employs local image feature matching~\cite{lowe1999object}, establishing correspondences between two images in a detector-based~\cite{luo2018geodesc,luo2020aslfeat,rublee2011orb,sarlin2020superglue} or detector-free ~\cite{li2020dual,liu2010sift,rocco2018neighbourhood,sun2021loftr} fashion. 
Other methods leverage point cloud registration algorithms~\cite{pomerleau2015review}, which detect 3D keypoints~\cite{bai2020d3feat,li2019usip}, extract feature descriptors~\cite{choy2019fully,deng2018ppfnet, gojcic2019perfect,hinterstoisser2016going,poiesi2021distinctive}, and estimate relative transformations~\cite{li2018deepim}. 
The recently proposed FS6D~\cite{he2022fs6d} further leverages complementary depth information in RGBD input images for dense prototype extraction and matching.
In comparison, our few-shot estimator leverages only RGB images, extending the applicable scenario to where depth information of objects is unavailable.
\vspace{-5mm}
\paragraph{Coarse-to-fine Estimation in 3D Vision.}
Coarse-to-fine estimation is a widely adopted strategy in 3D vision, where the estimation process is gradually refined in multiple stages, each with an increasing level of precision. ZebraPose~\cite{su2022zebrapose} proposes a coarse to fine training strategy to enable fine-grained correspondence prediction. Gen6D~\cite{gen6d} first finds the closest support view and then refines the initial pose using feature volume and 3D convolutional layers. Several coarse-to-fine multi-view stereo works~\cite{gu2020cascade,chen2019point,cheng2020deep,yang2020cost} estimate a coarse depth map and then gradually recover details based on feature pyramid. Neural Radiance Field (NeRFs)~\cite{nerf,barron2021mip} uniform sample along each ray to determine the rough density field and perform importance sample in the later stages.
\vspace{-5mm}
\paragraph{Self-supervised Feature Representation.}
Self-supervised learning is a well-studied topic in machine learning and computer vision~\cite{masci2011stacked,olshausen1996emergence,ranzato2007unsupervised,sims2016implementation}, with various methods for learning representations, such as clustering~\cite{bojanowski2017unsupervised,caron2018deep,ji2019invariant}, GANs~\cite{pathak2016context,mescheder2017adversarial}, pretext tasks~\cite{doersch2015unsupervised,noroozi2016unsupervised,wang2015unsupervised}, etc. 
For 3D vision, most methods of self-supervised learning focus on single object representation with different applications to reconstruction, classification or part segmentation~\cite{achlioptas2018learning,achituve2021self,gadelha2018multiresolution,groueix2018papier,hassani2019unsupervised,jing2020self,li2018so,sauder2019self}. 
Recent MVSFormer~\cite{cao2022mvsformer} proposes a pre-trained ViT-enhanced MVS network, which learns more reliable feature representations benefited by informative priors from self-supervised ViT.
Inspired by this work, in this paper, we study whether self-supervised ViT can facilitate feature learning in a similar task as 6D pose estimation.

\section{Methods}
\paragraph{Overview.}
Given $N_s$ RGB images (support views) of an object with known camera poses $\{\boldsymbol{I}^{(i)}, \boldsymbol{p}^{(i)} \}_{i=1}^{N_s}$, Cas6D predicts the object pose in the target view.
Cas6D first detects the object box in the target view, finds the top-$K$ best-matched viewpoints to initialize the $K$ poses, and conducts a coarse-to-fine refinement step using cascade feature volume and discrete pose hypotheses.

\subsection{Preliminary}

\paragraph{Generalizable 6DoF Estimation Framework}
In the model-free OnePose series~\cite{sun2022onepose,he2023onepose++}, object poses are determined by reconstructing a sparse point cloud from RGB sequences of support viewpoints and matching it with the 2D feature in the target view.
On the other hand, Gen6D~\cite{gen6d} breaks down the process into three steps. 1). \textbf{Detection:} the feature maps of support and target views are extracted using a VGG network~\cite{simonyan2014very}. Then, a detected box is obtained by convolving the feature maps, and a parallel branch predicts the box scale. 
2). \textbf{View Selection:} using the detected box, a view transformer estimates similarity scores and initial in-plane rotations to find the best-matched support view. 
3). \textbf{Pose Refinement:} 
a pose refiner will start from the pose of selected support view and iteratively adjust the 6DoF poses $ \boldsymbol{p} = [\boldsymbol{q}, \boldsymbol{t}, s] $ (parameterized by unit quaternion $ \boldsymbol{q} = [w, x, y, z] $, 3D translation vector $ \boldsymbol{t} = [t_x, t_y, t_z] $, and scale factor $s$) by estimating an update offset from a 32$^{3}$ volume built by unprojecting the 2D features.

Although a few generalizable 6DoF estimators~\cite{sun2022onepose,he2023onepose++,gen6d} have been investigated in the literature, they are known to be vulnerable in extreme sparse view settings, as the sparse point reconstruction may fail~\cite{sun2022onepose,he2023onepose++}, or the initial pose may be far from the ground truth due to false positive detections~\cite{gen6d} (see Figure~\ref{fig:teaser}). Furthermore, Gen6D~\cite{gen6d} suffers from GPU memory issue that limit its method to low feature and voxel resolutions (32$^2$ and 32$^3$, respectively). On the other hand, OnePose series~\cite{sun2022onepose,he2023onepose++} requires an additional YOLO detector, which may limit its deployment for out-of-distribution objects. 
To overcome the challenges posed by sparse view settings, we propose three improvements upon the three stages in the pipeline of Gen6D, respectively: robust object detection, parallel pose initialization, and cascade feature volume pose refinement.
These tailored designs for few-view settings altogether significantly improve accuracy and robustness in the extremely sparse view scenarios.

\subsection{Cascade Refinement using Feature Pyramid}
Among the three stages, pose refinement plays a crucial role in accurate pose regression.
Overall, the iteratively refined general 6DoF pose estimation can be formulated as an optimization problem toward the objective below:
\vspace{-2mm}
\begin{align} \label{eqn:obj}
    \mathcal{L}(\boldsymbol{p}) = \sum_{i=1}^{N_s} \ell(\mathcal{W}(\boldsymbol{I}^{(i)} \vert \boldsymbol{p}^{(i)}, \boldsymbol{p}), \boldsymbol{I}),
\end{align}
where $\mathcal{W}$ wraps the $i_{th}$ support(reference) images $\boldsymbol{I}^{(i)}$ into the tentative target pose $\boldsymbol{p}$, by using the $i_{th}$ support(reference) pose $\boldsymbol{p}^{(i)}$. $\ell$ computes a general matching loss over two frames.
Both $\mathcal{W}$ and $\ell$ can be applied to either key points or whole images~\cite{sun2021loftr}.
To optimize variable $\boldsymbol{p}$, gradient-based methods are typically adopted, e.g., we can iterative descent the error $\mathcal{L}$ by $\boldsymbol{p}_{t+1} = \boldsymbol{p}_{t} - \eta \nabla \mathcal{L}(\boldsymbol{p}_{t})$.
However, since pose estimation faces severe ambiguity, it is intractable to consider a general closed-form $\mathcal{W}$ and $\ell$. Also, the resultant Eq. \ref{eqn:obj} is highly non-convex, which causes the optimization prone to terminate at a saddle but inaccurate point.
The seminal work Gen6D \cite{gen6d} proposes to surrogate the gradient term with a data-driven trained neural network $s_{\theta}$, which directly predicts the update in the pose space:
\vspace{-2mm}
\begin{align} \label{eqn:refiner}
    \boldsymbol{p}_{t+1} = \boldsymbol{p}_{t} - \eta s_{\theta} \left(\boldsymbol{p}_{t} \left| \boldsymbol{I}, \left\{ \boldsymbol{I}^{(i)}, \boldsymbol{p}^{(i)} \right\}_{i=1}^{N_s} \right. \right).
\end{align}
We note that the learned $s_{\theta}$ involves the implicit modeling of both $\ell$ and $\mathcal{W}$, which is more flexible and adaptive to real-world scenarios without handcrafted bias.
As illustrated in Eq. \ref{eqn:refiner}, the neural network needs to aggregate support view images and evaluate the matching score with the pose at the current iteration.
Below we introduce our architectural design for $s_{\theta}$ to improve its robustness under few-view setting.

We first obtain the detected object box in the full picture and select its initialized pose $\boldsymbol{p}_0$ (also known as the pose in the nearest support view).%
With the proper initialization, the pose refiner $s_{\theta}$ should estimate the update according to the current pose and reference images.
To implement the pose refiner, we create a volume within the unit cube at the origin with 32$^3$ vertices. The volume features are derived from two sources: 1) the unprojected feature of the six nearest support images, and 2) the unprojected feature with initial poses and the object in the target view, with mean and variance computed across all support views.
The constructed volume will then pass through a 3D ConvNet to be mapped to an update vector in the pose space.

In Gen6D \cite{gen6d}, only a single-scale volume is constructed.
While a lower-grained volume tentatively mitigates a large pose gap, a finer-grained volume carries more details and improves estimation accuracy. Therefore, the single-scale volume is not optimal under the sparse view setting.
As discussed before, the pose refiner predicts a gradient to minimize the matching score between target and support views on the volume space \cite{gen6d}.
However, the matching procedure is sensitive to fine details especially when observations are limited, likely to overfit to the high-frequency noises 
\cite{lowe2004distinctive, shi1994good}.
To conquer these challenges, we propose a cascade volume feature formulation and adopt a coarse-to-fine prediction approach.
The rationale is that we start from a coarse-grained volume, which provided low-frequency and contour information to align the rough orientation and position, then we gradually increase the granularity to inject high-frequency signals, which fine-tunes the pose locally to match the low-level details.
\vspace{-5mm}
\paragraph{Feature Pyramid Volume Construction.} To strike a balance between volume resolution and computational efficiency, prior work \cite{gen6d}, constructed a volume with $32^3$ vertices, each containing 128 feature dimensions. However, such a single-scale (top-level) feature representation and $32^3$ volume spatial resolution are not able to accurately regress poses. In contrast, we leverage the Feature Pyramid Network~\cite{lin2017feature} and utilize its feature maps with progressively increased spatial resolutions to construct feature volumes of higher resolutions. Specifically, we construct a three-stage volume with spatial resolutions of \{$16^3$, $32^3$, $64^3$\} by extracting feature maps \{P2, P3, P4\}. The feature dimensions of these volumes, obtained through the process of unprojecting from FPN, are \{64, 32, 16\}.
\vspace{-5mm}

\begin{table*}[ht!]
\begin{center}
\resizebox{0.7\linewidth}{!}{
\begin{tabular}{lllllllllll}
\toprule[1pt]
\multirow{2}{*}{\textbf{Type}}& \multirow{2}{*}{\textbf{Method}} & \multicolumn{8}{c}{\textbf{Object Name}}                                    & \multirow{2}{*}{\textbf{Avg.}} \\ 
\cline{3-10}
&& cat & duck & bvise & cam & driller  & lamp & eggbox$^*$  & glue$^*$      &      \\ \midrule
&                & \multicolumn{8}{c}{\textbf{ADD(S)-0.1d}}                                                         &       \\ \midrule
\multirow{5}{*}{\textbf{16-shot}}   
& CDPN \cite{li2019cdpn}    & 0.00 &1.20 &0.00 &0.10 &0.66 &0.25 &0.00 &4.25 &0.81 \\
& EPro-PNP\cite{chen2022epro} & 0.10 &0.00 &0.00 &0.10 &0.79 &0.38 &0.00 &5.21 &0.82\\ 
& Gen6D~\cite{gen6d}   & \underline{23.55}   & \underline{15.96}    & 41.76      & \underline{28.30}  & 35.67  &  60.74       &  \underline{92.39}   &  \underline{58.60} &  \underline{44.62}     \\ 
& OnePose++~\cite{he2023onepose++} & 20.00 & 5.07  & \textbf{57.61}& 23.43   & \textbf{50.80}  & \underline{63.60}  & 81.40   & 28.50  &  41.30    \\ 
 & Ours &\textbf{27.85} & \textbf{19.25} &\underline{45.83} &\textbf{29.7} & \underline{39.54} & \textbf{63.84} & \textbf{93.95} & \textbf{62.64} &\textbf{47.83 }         \\ \hline
\multirow{5}{*}{\textbf{32-shot}}
& CDPN \cite{li2019cdpn} & 2.69 & 2.54 & 17.17 & 5.88 & 7.14 & 11.92 & 11.97 &  7.01 & 8.29 \\  
& EPro-PNP\cite{chen2022epro} & 1.80 & 1.50 & 21.24 & 6.27 & 15.06 & 12.09 & 19.81 & 10.71 & 11.06\\ 

& Gen6D~\cite{gen6d} & \underline{44.81}& \underline{32.39}     & 62.30    & 51.56    & 56.10 & 81.76   &\textbf{98.21}   &  \textbf{76.64} & \underline{62.97}          \\
& OnePose++~\cite{he2023onepose++}    & 35.42   & 24.31     & \textbf{72.55}     & \textbf{71.07}  & \textbf{71.55} &    \textbf{85.31}  &   95.68         &   32.91 & 61.10       \\  
& Ours  & \textbf{48.55} & \textbf{40.66} & \underline{65.98} &\underline{52.68} & \underline{61.64} & \underline{82.53}&\textbf{98.21}& \textbf{76.64} & \textbf{65.86}       \\ \hline
\multirow{5}{*}{\textbf{Full-shot}} 
& CDPN \cite{li2019cdpn} &  \underline{86.63} & \underline{75.21} & \textbf{98.74} & \underline{92.84} & \underline{95.14} & 97.79 & 99.62 & \textbf{99.61} & \underline{93.20} \\
& EPro-PNP\cite{chen2022epro} & \textbf{90.52} & \textbf{80.09} & 96.41 & \textbf{92.94} & \textbf{95.94} & \textbf{99.04} & \textbf{99.91} & \underline{98.17} & \textbf{94.13}   \\ 
& Gen6D \cite{gen6d}  & 60.38    & 40.28    & 77.13   & 66.07 &    67.09 &  89.80   &   98.31&  87.83 & 73.30                 \\  
& OnePose++~\cite{he2023onepose++}    & 70.40 &42.30 & \underline{97.30} &88.00 &92.50 & \underline{97.80} & \underline{99.70} &48.00 &79.50\\ 
& Ours           & 60.58 &51.27 &86.72 &70.10 &84.84 &93.38 &98.78 &88.51 &79.27                 \\ \midrule
&                         
& \multicolumn{8}{c}{\textbf{Prj-5}}        &                       \\ \midrule
\multirow{5}{*}{\textbf{16-shot}}  
& CDPN \cite{li2019cdpn} & 1.20 &5.88 &0.00 &0.00 &2.00 &1.25 &1.50 &1.25 &1.63  \\   
& EPro-PNP\cite{chen2022epro} &1.30 & 1.88 & 0.97 & 2.65 & 3.07 & 1.63 & 1.50 & 2.61 & 1.95\\   

& Gen6D~\cite{gen6d}       & \underline{68.46}    & \underline{61.12}     & 41.08      & \underline{46.40}   & 31.70    &    49.90    &   \underline{88.07}  &  \underline{77.4} &  \underline{58.03}         \\ 
& OnePose++~\cite{he2023onepose++}     & 37.25   & 14.93     & \textbf{62.17}      & 37.25    & \textbf{58.57}  &   \textbf{72.93} &     49.76    &   24.40
&  44.66            \\   
& Ours     & \textbf{76.45} &\textbf{75.49} &\underline{46.28} & \textbf{57.23} & \underline{48.40} & \underline{53.67} & \textbf{90.07} & \textbf{82.59} & \textbf{66.27} \\ \midrule
\multirow{5}{*}{\textbf{32-shot}}   
& CDPN \cite{li2019cdpn} & 12.08 & 47.70 &  14.94 & 13.43 & 4.32 & 3.24 & 7.32 & 11.20  & 14.28 \\   
& EPro-PNP\cite{chen2022epro} & 11.18 & 38.31 & 19.59 & 15.39 & 9.91 & 5.85 & 14.46 & 16.70 & 16.42  \\   
& Gen6D~\cite{gen6d}  & \underline{92.71}    & \underline{76.61}   & 67.83   & 84.11    & 63.13     & 80.42   & \textbf{95.96}  &    \underline{93.91}    &   \underline{81.84} \\   
& OnePose++~\cite{he2023onepose++}    & 83.63    & 69.95     & \textbf{81.37}      & \textbf{94.40}    & \textbf{72.94}       &  \textbf{90.59 }&  89.10  & 30.41 & 76.50    \\   
& Ours            & \textbf{94.01} & \textbf{92.39} & \underline{69.79} & \underline{85.03} & \underline{69.17} & \underline{84.64} & \underline{95.58} & \textbf{95.24} & \textbf{85.70}     \\ \midrule
\multirow{5}{*}{\textbf{Full-shot}}
& CDPN \cite{li2019cdpn} & \underline{99.30} & \underline{98.40} & \underline{98.74} &  98.74 & 94.85 & 95.68 & \underline{99.06} & 98.36 & \underline{97.89} \\   
& EPro-PNP\cite{chen2022epro} & \textbf{99.80} & \textbf{99.06} &98.64& \underline{99.12} & \textbf{97.32} & \underline{97.02} & \textbf{99.25} & \textbf{98.94} & \textbf{98.64} \\   
& Gen6D~\cite{gen6d} & 96.10    & 79.71    &  82.46      & 90.78    &72.44  & 91.60 &  97.84    &  96.23 &  88.40  \\   
& OnePose++~\cite{he2023onepose++} & 98.70    & 97.70     & \textbf{99.60}      & \textbf{99.60}    & 93.10    &  \textbf{98.80}     &    98.70     & 51.80 & 94.25       \\   
& Ours         & 99.00 &93.50 &93.41 &96.27 & \underline{94.95} & 96.93 &98.31 & \underline{98.84} &96.40                       \\ \bottomrule
    \end{tabular}
}
\end{center}
\caption{\textbf{Quantitative Comparison on LINEMOD Dataset.} The table shows the performance comparison of Cas6D with several generalizable methods in both sparse view and full-shot settings. The best and second-best performing methods are highlighted in bold and underlined, respectively. Cas6D outperforms previous generalizable (model-free) methods consistently in all settings.}
\vspace{-5mm}
\label{tab:main_limod}
\end{table*}

\paragraph{Pose Hypothesis Range in Each Stage}
To improve the accuracy of initial pose estimation using the volume feature, there are two approaches: manually setting the sampling range $\mathcal{R}_t$ at each stage $t$, or caching the distance between predictions and ground truth over the entire training set. Subsequently, the sampling range can be gradually reduced in each stage by a scaling factor $w_t$, such that $\mathcal{R}_{t+1}$ = $w_t \times \mathcal{R}_t$. In our implementation, we choose $w_t < 1$ to progressively narrow down the pose search range.
\vspace{-5mm}
\paragraph{Camera Pose Parameterization}
To refine the estimated 6DoF pose $\boldsymbol{p}_t$ in the $t$-th stage, we can perform direct regression of the residual using the Mean Squared Error (MSE) loss. However, under sparse view scenarios, where there is a large gap between initializations and ground truth, this approach may lead to unstable learning. To address this issue, we discretize the measured pose range into a set of discrete bins and use classification loss for pose bin estimation. We denote each interval of adjacent bins as $I_i$ in stage $i$. The pose estimation task then becomes a classification problem of assigning the correct viewspace bin to the initial poses. In the later stages, we apply finer pose intervals to recover more detailed poses: $I_{t+1} = v_t \times I_t$, where $v_t < 1$ narrows the bin interval gradually.

\subsection{Parallel Pose Initialization}
In the case of sparse input, performance is significantly affected by the inaccurate pose initialization when searching for the closest support images. Previous work, such as FS6D, assumes that accurate depth maps are available for all captures. However, depth maps are not always available for everyday objects. To mitigate the degradation of performance in the sparse-view setting, we propose an initialization scheme that generates multiple pose hypotheses based on the top-K similarity scores between the target view and all support views. With this initialization scheme, the refiner can obtain a better starting point by warping the support to the target feature volumes, as some of the hypotheses may be relatively close to the ground truth.

\subsection{Detection with Self-supervised Semantic Correspondence.}
One of the most common failure modes in 6DoF estimation is the inability of the network to find feature correspondences between different views of the same object, which can be caused by noisy backgrounds or low-textured objects. This problem is further amplified in the few-shot setting where fewer observations are available. To address this issue, we draw inspiration from the 2D co-segmentation work~\cite{amir2021deep} which demonstrates that DINO, a powerful feature extractor, can learn feature correspondences for image pairs. Therefore, we incorporate the frozen DINO-ViT~\cite{caron2021emerging} feature to aid the detector in localizing the correct object box. The adoption of the frozen DINO feature incurs only negligible memory cost, which will be discussed in the experiment section.

\subsection{Loss Function}
The three modules are trained independently with their own losses. We describe the loss function of the cascade refiner as follows and state other losses in the supplementary due to space limitation.
We first transform the quaternion representation to rotation matrix and translation vector, sample $v_i^3$ voxel points in each stage of the object coordinate system ($v_i$ is voxel resolution), and transform them into camera coordinate system using object poses. We calculate the distance of the transformed points by using predicted and the ground-truth poses.
\vspace{-3mm}
\begin{equation}
    \ell_{pose}=\sum_{k=1}^{N} \lambda_{k}\cdot\sum_{m=1}^{v_{k}^{3}} \left\| \mathcal{T}_{pr}(p_k\mid k,m ) - \mathcal{T}_{gt}(p_k\mid k,m ) \right\|_{2}
\end{equation}
where $k$ means refine stage index, $m$ is voxel index, $\lambda_k$ indicate loss weights of each stage, and $\mathcal{T}_{pr}(p_k\mid k,m)$ and $\mathcal{T}_{gt}(p_k\mid k,m )$ indicate the transformed points, respectively.

\section{Experiment}

\subsection{Experiment Setup}
\paragraph{LINEMOD Dataset~\cite{hinterstoisser2013model}.} The LINEMOD dataset is a widely used dataset for object pose estimation. It includes 13 videos, each featuring 13 low-textured objects, and approximately 1000 test images for each object, which are used as target images for evaluation.

\vspace{-5mm}
\paragraph{GenMOP Dataset~\cite{gen6d}.} The GenMOP dataset~\cite{gen6d} comprises 10 objects captured in two video sequences under varying environmental conditions, such as different backgrounds and lighting conditions. Each video sequence is composed of approximately 200 images.

\vspace{-5mm}
\paragraph{Training Details}
We conducted experiments with other generalizable methods~\cite{gen6d}  using the same training datasets, which include 2000 ShapeNet models~\cite{chang2015shapenet}, 1023 objects from the Google Scanned Object dataset~\cite{wang2021ibrnet}, 5 objects from the GenMOP dataset~\cite{gen6d}, and 5 objects from the LINEMOD dataset~\cite{hinterstoisser2013model}.
To evaluate the sparse-view scenario, we randomly selected 16 to 64 views for training the refiner of our method and Gen6D~\cite{gen6d}.
We select the top-3 pose candidates as initialization and evaluate the comparisons using Cas6D with 3-stage with volume feature size as ({16$^2\times$64, 32$^2\times$32, 64$^2\times$16}). We cache and update the maximum pose difference to determine the range.
Throughout all stages, we set annealing factors $v_i$ and $w_i$ to 0.5, pose bin number as \{16, 8, 4\} for all components in 6DoF poses $\boldsymbol{p}$.
We use the same data pre-processing and training recipe as Gen6D~\cite{gen6d}.
We select the sparse views from all evaluation views for evaluation using farthest
point sampling (FPS)~\cite{gen6d}.
All experiments were performed using 8 Tesla V100 GPUs with 32GB memory.

\vspace{-5mm}
\paragraph{Metrics}
We adopted the commonly used \textit{Average Distance}(ADD) and \textit{Projection Error} as metrics to evaluate the performance of all methods. For an object $\boldsymbol{\mathcal{O}}$ composed of vertices $\boldsymbol{v}$ , the ADD/ADDS for both asymmetric and symmetric objects with the predicted pose $\mathbf{R}$, $\mathbf{t}$ and ground truth pose $\mathbf{R^{*}}$ , $\mathbf{t^{*}}$ were calculated using the following formulas: 
\vspace{-3mm}
\begin{align}
\mathrm{ADD}&=\frac{1}{m} \sum_{v \in \mathcal{O}}\left\|(\mathbf{R} v+\mathbf{t})-\left(\mathbf{R^{*}} v+\mathbf{t^{*}}\right)\right\| \\
\mathrm{ADDS}&=\frac{1}{m} \sum_{v_{1} \in \mathcal{O}} \min _{v_{2} \in \mathcal{O}}\left\|\left(\mathbf{R} v_{1}+\mathbf{t}\right)-\left(\mathbf{R^{*}} v_{2}+\mathbf{t^{*}}\right)\right\| 
\end{align}
\vspace{-1mm}
We computed the recall rate on the ADD and ADDS metrics with 10\% of the object diameter (ADD-0.1d and ADDS-0.1d).
For the projection error, we computed the recall rate at 5 pixels and referred to it as Prj-5.

\subsection{Comparisons}

\begin{figure*}[!t]
    \centering
    \includegraphics[width=0.9\linewidth]{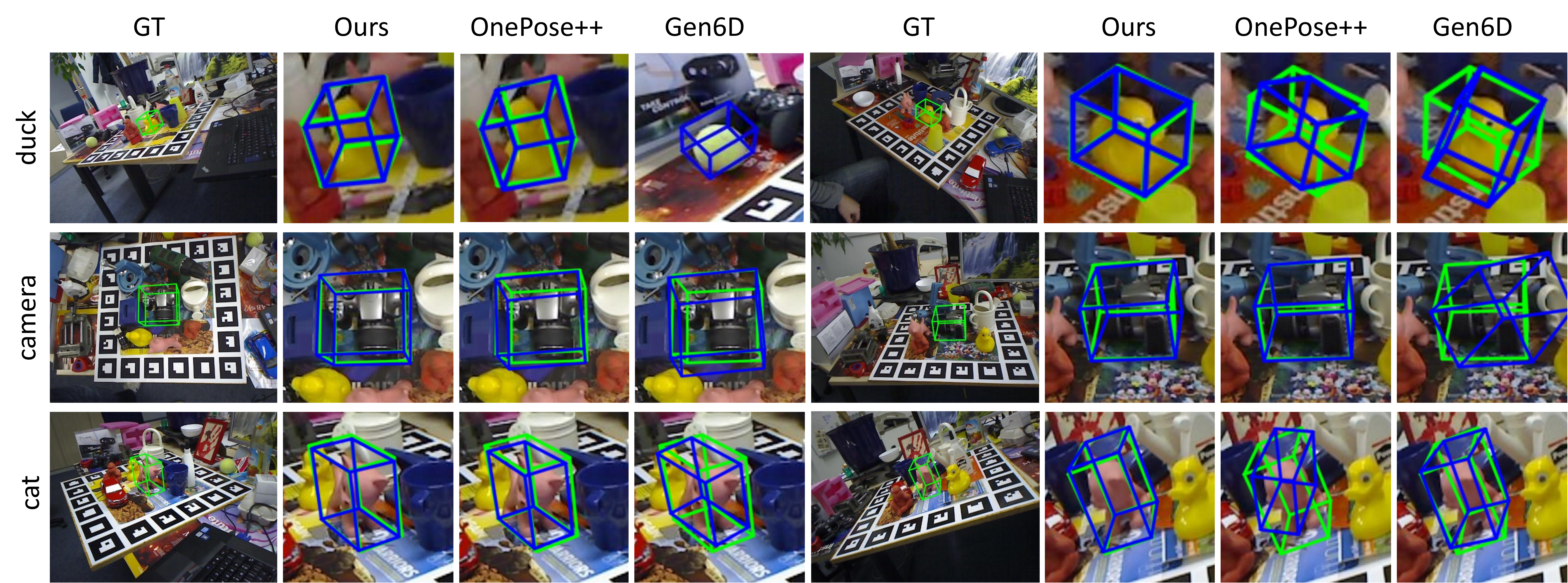} \vspace{-0.3em}
    \caption{\textbf{Qualitative results on the LINEMOD dataset under 32-shot setting.} The ground-truth poses are drawn in \textcolor{green}{green} while estimated poses are drawn in \textcolor{blue}{blue}. Gen6D~\cite{gen6d} may erroneously detect the incorrect object with a similar appearance, while Cas6D is more robust against appearance noises. Further, Cas6D usually generates a more tight 3D orientated box than other methods, owning to the coarse-to-fine refinement by using 2D feature pyramid.
}\label{fig:linemod_main}
\vspace{-3mm}
\end{figure*}

\begin{table}[]
\resizebox{0.99\linewidth}{!}{
\begin{tabular}{llllllll}\toprule
\multirow{2}{*}{\textbf{Type}}      & \multirow{2}{*}{\textbf{Method}} & \multicolumn{5}{c}{\textbf{Object Name}}                                                                                                & \multirow{2}{*}{\textbf{Avg.}} \\ \cline{3-7}
                           &                         & \multicolumn{1}{l}{Chair} & \multicolumn{1}{l}{PlugEN} & \multicolumn{1}{l}{Piggy} & \multicolumn{1}{l}{Scissors} & TFormer &                       \\ \midrule
                           &                         & \multicolumn{5}{c}{\textbf{ADD(S)-0.1d}}                                                                                                &                       \\ \midrule
\multirow{3}{*}{\textbf{16-shot}}   & Gen6D   & \underline{48.50} &\underline{10.54} &\underline{59.80} &24.57 &\underline{26.59} & \underline{34.00}        \\    
                           & OnePose++ &   25.00 &1.87 &47.24 &\underline{25.43} &21.42 &24.19   \\    
                           & Ours      &   \textbf{52.87} &\textbf{12.58} &\textbf{64.09} &\textbf{26.80} &\textbf{28.74} & \textbf{37.01}                     \\ \hline
\multirow{3}{*}{\textbf{32-shot}}   & Gen6D  & \underline{51.50} &\underline{11.68} &65.32 &32.75 &\underline{44.0} & \underline{41.05}   \\   
                                 & OnePose++ &   31.0 &4.67 &\textbf{70.85} &\textbf{36.64} &26.19 &33.87       \\    
                           & Ours   & \textbf{54.85} &\textbf{15.45} &\underline{67.94} &\underline{35.17} &\textbf{46.61} &\textbf{44.00}                     \\ \hline
\multirow{3}{*}{\textbf{Full-shot}} & Gen6D    &\underline{61.50} &\underline{19.63} &74.37 &33.62 &\underline{63.89} &50.60                \\                                 
                           & OnePose++  &   58.50 & 4.67 & \textbf{94.47} &  \textbf{39.22} &  59.13 & \underline{51.20}                   \\    
                           & Ours                &\textbf{64.00} &\textbf{23.36} &\underline{77.87} &\underline{36.88} & \textbf{65.47} &\textbf{53.52  }                   \\ \midrule
                           &   & \multicolumn{5}{c}{\textbf{Prj-5}}                                                    &                       \\ \midrule
\multirow{3}{*}{\textbf{16-shot}}   & Gen6D   & 33.50 & \underline{66.43} & \underline{90.45} &\underline{79.31} & \underline{90.87} & \underline{72.11}   
                            \\    
                           & OnePose++   &  \textbf{40.00} &11.21 &19.6 &44.83 &63.10 &35.75         \\    
                           & Ours  &   \underline{39.58} & \textbf{73.11} &\textbf{95.49} &\textbf{84.74} &\textbf{93.50} &\textbf{77.28 }          \\ \hline
\multirow{3}{*}{\textbf{32-shot}}   & Gen6D   & 42.50 & \underline{68.69} & \underline{94.98} & \underline{87.06} & \underline{91.50} & \underline{76.95}      \\    
                           & OnePose++  &  \textbf{49.50} &14.48 &56.28 &70.26 &79.36 &53.98                \\    
                           & Ours    &  \underline{46.71} &\textbf{86.41} &\textbf{99.60} & \textbf{90.83} &\textbf{99.50} &\textbf{84.61}                    \\ \hline
\multirow{3}{*}{\textbf{Full-shot}} & Gen6D  &46.71 &70.41 &\underline{98.60} & \underline{96.83} & \underline{96.37} & 81.78           \\    
                          & OnePose++ & \textbf{81.00}  &  \underline{75.23} &  95.98 & 82.75 &  90.48 & \underline{85.09}  \\    
                           & Ours     & \underline{62.50} &\textbf{78.67} & \textbf{99.50} &\textbf{97.43} &\textbf{99.50} &\textbf{87.52}                  \\    
                         \bottomrule\\
\end{tabular}}
\caption{\textbf{Quantitative results on GenMOP dataset.} We report the comparisons with several generalizable (CAD model-free) methods, under sparse view settings and full-shot settings.}
\vspace{-3mm}
\label{tab:main_genmop}
\end{table}
We compared the proposed Cas6D with two categories of baselines. The first category is generalizable methods~\cite{gen6d,he2023onepose++} with the same training datasets as ours. All objects for evaluation are not seen in the training sets.
Note that, OnePose++~\cite{he2023onepose++} leverage an additional YOLOv5~\footnote{\url{https://github.com/ultralytics/yolov5}} as object detector.
The second category is instance-level methods~\cite{li2019cdpn,chen2022epro}  that require CAD models and are separately trained for each object in the test set. 

\paragraph{Results on LINEMOD}
We compare the proposed method with other state-of-the-art methods under \{16, 32, full(all support)\} views settings. 
Specifically, we re-trained the refiner of the original Gen6D~\cite{gen6d} model, which randomly selects from 16 to 64 support views to find the nearest support to align with our training recipe. 
As is shown in Table~\ref{tab:main_limod}, Cas6D consistently outperforms other generalizable baselines under all settings. Under the full-shot setting, instance-level methods perform with almost 100\% accuracy owing to their use of ground-truth CAD models.
The performance difference between our proposed method and OnePose++\cite{he2023onepose++} can be attributed to the latter's inability to reconstruct an accurate point cloud when the number of available support views decreases, making it difficult to find the best matches. Figure~\ref{fig:linemod_main} illustrates the qualitative results, where it can be observed that Gen6D fails to detect the correct object box, leading to an incorrect pose regression result (4th column). In contrast, our Cas6D is more robust under the sparse-view setting, enabling it to detect the object pose accurately and estimate the final object pose.

\begin{figure*}[!t]
    \centering
    \includegraphics[width=0.9\linewidth]{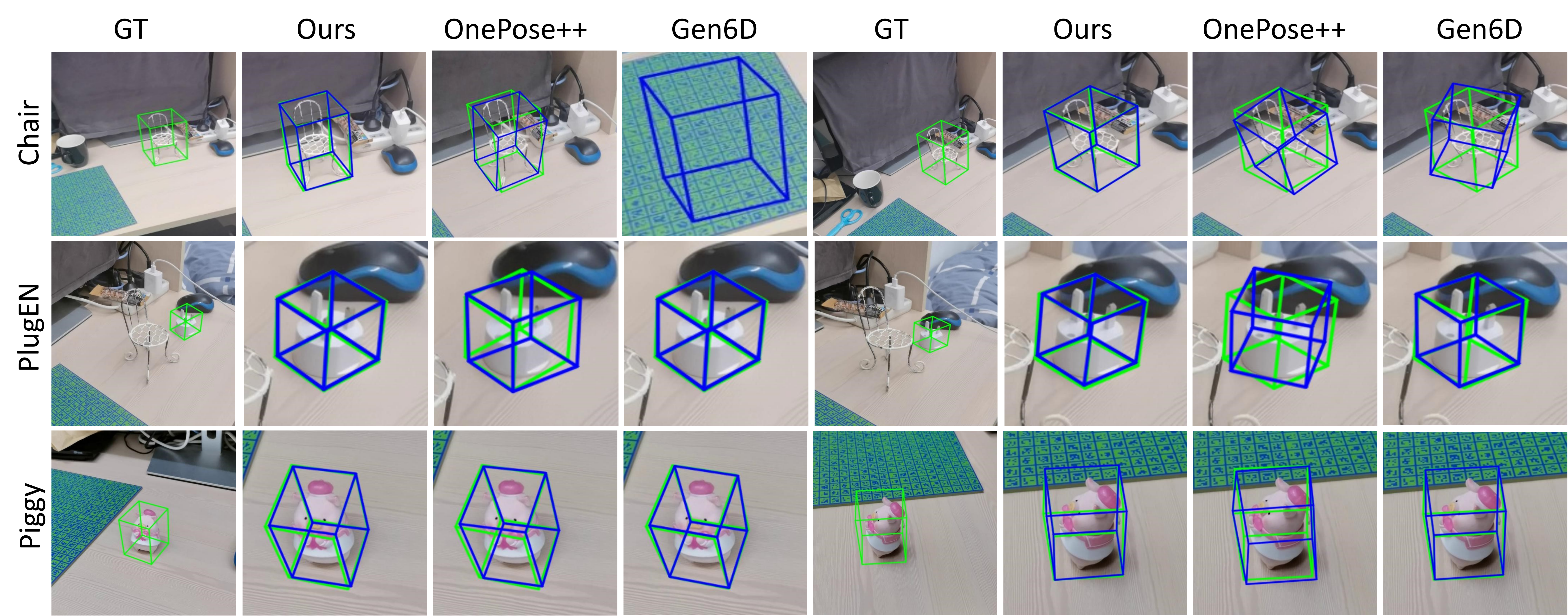} 
    \vspace{-1mm}
     \caption{\textbf{Qualitative results on the GenMOP dataset~\cite{gen6d} under 32-shot setting.} The ground-truth poses are drawn in \textcolor{green}{green} while estimated poses are drawn in \textcolor{blue}{blue}. Cas6D performs consistently better than other baselines, by utilizing the robust object detection, multiple-pose initialization and coarse-to-fine pose refinement.
     \vspace{-3mm}
}\label{fig:main_genmop}
\end{figure*}

\begin{table*}[t]
\begin{center}
\vspace{3mm}
\resizebox{0.98\linewidth}{!}{
\begin{tabular}{ccccccccc}
 \toprule 
ID & DINO Feat.? & Feature Pyramid?   & Init. Pose Num. & Stage Num. & Reg./Cls.?          & Avg. ADD-0.1d & Avg. Prj-5 & GPU Mem. \\ \midrule
1 &  \redtime  & \redtime   & 1  & 1 & Reg &63.00   & 81.80     &826M      \\ 
2 & \redtime   & \greencheck   & 1  & 1 & Reg &63.20   &82.15    &846M      \\ 
3 & \greencheck   & \greencheck   & 1  & 1 & Reg &63.80  & 82.90   &886M      \\ 
4 & \greencheck   & \greencheck   & 1  & 2 & Cls & 65.20 &84.00& 695M\\ 
5 & \greencheck   & \greencheck   & 1  & 3 & Cls &65.50 &84.80  & 789M  \\ 
6 & \greencheck   & \greencheck  & 3  & 3  & Cls &65.86 &85.70  & 805M   \\ 
7 & \greencheck   & \greencheck  & 3  & 3  & Reg &63.68 & 83.29  &805M\\
8 & \greencheck   & \greencheck  & 5  & 3  & Cls & 65.28 &84.83  &815M \\ 
  9 & \greencheck   & \greencheck  & 7  & 3 & Cls & 63.69 &81.43  &835M \\  \bottomrule
  \\
\end{tabular}}
\caption{\textbf{Ablation Studies.} We perform extensive experiments to validate our proposed designs, by reporting the pose estimation accuracy, and GPU memory consumption at the inference stage. Results are averaged on LINEMOD~\cite{hinterstoisser2013model} dataset under the 32-shot setting.}\label{tab:ablation}
\vspace{-6mm}
\end{center}
\end{table*}

\vspace{-5mm}
\paragraph{Results on GenMOP}
We evaluate the generalization ability of our proposed framework on the GenMOP dataset~\cite{gen6d} and use the same evaluation criterion as that of the LINEMOD dataset. Table~\ref{tab:main_limod} presents the comparison results, and we find that Cas6D outperforms the compared baselines under sparse-view settings.
Furthermore, the visualized pose estimations on various objects in Figure~\ref{fig:main_genmop} further validate the effectiveness of our method, which can be attributed to the robust object detection, parallel pose initialization, and cascade pose refinement with feature pyramid.

\subsection{Ablation Studies}
We conduct ablation studies on the LINEMOD test set~\cite{hinterstoisser2013model} to verify the effectiveness of the proposed design. The results are presented in Table~\ref{tab:ablation}.
\vspace{-5mm}
\paragraph{Volume Construction with Feature Pyramids}
In order to demonstrate the advantage of volume feature construction using feature pyramids, we compare our approach with single-scale construction~\cite{gen6d} in this ablation study. We replace the single scale feature in Gen6D~\cite{gen6d} (Table~\ref{tab:ablation}, ID 1) with our feature pyramids (Table~\ref{tab:ablation}, ID 2) while keeping the single-scale volume feature size fixed at 32$^3 \times$ 128. We found that using feature pyramids led to an improvement in averaged ADD-0.1d/Prj-5 from 63.0/81.8 to 63.2/82.15, with negligible GPU memory overheads. This validates that fusing multi-scale features enriches the volume representation and leads to better results.

\vspace{-5mm}
\paragraph{How Many Cascade Stages Should be Used?}
We conduct experiments with different cascade stage numbers ({2, 3}) and report the corresponding results in Table~\ref{tab:ablation} (ID 4 and 5). We only use one pose candidate as the input of the cascade refiner for this comparison. The volume feature is constructed using FPN feature maps from three spatial levels ({16$^2$, 32$^2$, 64$^2$}) with volume sizes of ({16$^2\times$64, 32$^2\times$32, 64$^2\times$16}). Our results show that as the number of stages increases, the performance first improves and then stabilizes.

\begin{figure}[]
\begin{center}
\vspace{-2mm}
\includegraphics[width=1\linewidth]{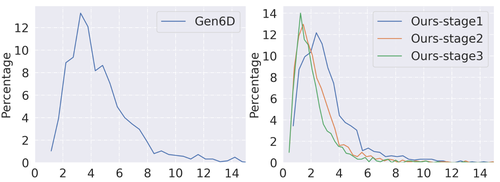}
\end{center}
\vspace{-2mm}
\caption{\textbf{Averaged Pose Residual:} We show the percentage (y-axis) and the pose error from GT (x-axis).
\vspace{-2mm}
}
\label{fig:range}
\vspace{-2mm}
\end{figure}

\vspace{-5mm}
\paragraph{The Effect of Coarse-to-fine framework}
Our coarse-to-fine framework with a \textit{novel pose parameterization} and \textit{adaptive pose sampling} addresses the limitation of a regression-based single-scale pose volume feature (Figure~\ref{fig:range}, left), enabling a wider pose search range and fine-grained predictions (Figure~\ref{fig:range}, right).

\vspace{-5mm}
\paragraph{Impact of Parallel Pose Initialization}
The use of multiple parallel pose initializations from the similarity network helps the refiner to find the support view with the lowest feature warping error. To test this, we designed experiments where we used the top \{1, 3, 5, 7\} pose initializations with the highest similarity score. Table~\ref{tab:ablation} shows the results in IDs 5, 6, 8, and 9. The accuracy increases when more potential support views are included, but when more than 5 views are used, the performance starts to degrade because these pose candidates may be too far from the target view.

\vspace{-5mm}
\paragraph{Shall We use Regress or Discrete Pose Bins?}
We conducted a comparison between regression and classification-based pose residual estimation methods under sparse view settings. As shown in ID 6 and 7 of Table~\ref{tab:ablation}, we observe that our method performed better when mapping the gap (from initial to GT) to discrete pose bins, achieving 85.70 on Prj-5, compared to 83.29 achieved by the regression-based method. This indicates that using discrete pose bins is more effective in sparse view scenarios.

\begin{table}[]
\begin{center}
\vspace{-3mm}
\resizebox{\linewidth}{!}{
\begin{tabular}{lllllll}
\toprule \text {  } & \text { cat } & \text { duck } & \text { bvise } & \text { cam } & \text { driller } & \text { Avg. } \\
\midrule \text { Gen6D } & 76.99 & 42.15 & 63.33 & 72.92 & 48.78 & 60.83 \\
\text { Cas6D(Ours) } & 79.46 & 67.44 & 66.32 & 76.39 & 59.35 & 69.70 \\
\bottomrule
\end{tabular}
}
\end{center}
\caption{\textbf{Box Detection Accuracy.} We evaluated the object detection accuracy on the LINEMOD dataset under the 32-shot setting. Cas6D outperforms Gen6D in terms of mAP@.5:.95 (\%) on the test set, demonstrating that the self-supervised feature representation benefits correspondence matching across object views.}\label{tab:abalation_iou}
\vspace{-6mm}
\end{table}

\vspace{-6mm}
\paragraph{The Effect of Pre-trained ViTs}
The effectiveness of self-supervised trained representation has been demonstrated in 2D co-segmentation~\cite{amir2021deep}. In this work, we demonstrate the importance of pre-training for feature learning in the entire 6DoF pipeline for object detection. The average ADD-0.1d/Prj-5 improves from 63.2/82.15 to 63.80/82.90 with negligible impact on GPU memory overheads (IDs 2 and 3 of Table~\ref{tab:ablation}). We also evaluate the mAP@.5:.95 (\%) of the detected box against the ground truth and observe an improvement from 60.83 to 69.70, as presented in Table~\ref{tab:abalation_iou}.

\section{Conclusion, Limitations and Future Works}
In this paper, we propose a cascade 6DoF pose estimation framework, called \textbf{Cas6D}, for few-shot object pose estimation with multi-view RGB images. Our approach leverages self-supervised pre-trained DINO-ViT representations and a divide-and-conquer model design to formulate a robust arbitrary object detector. We use top-K similar support views to initialize multiple pose candidates and design a cascade volume feature formulation to generate finer poses with finer volume features, narrowing the pose search range at each stage using multi-scale FPN features. Our framework outperforms all other generalizable baselines under both few-shot and full-shot settings while reducing inference memory requirements.
However, the current implementation of our framework requires constructing 3D volumes and performing 3D convolutional layers for pose regression, limiting its application to ultra-high resolution scenarios. We plan to explore more efficient implementations of volume feature aggregation.

{\small
\bibliographystyle{ieee_fullname}
\bibliography{egbib}
}

\end{document}